\documentclass[11pt,a4paper]{article}
\usepackage[hyperref]{acl2020}
\usepackage{times}
\usepackage{latexsym}

\usepackage{microtype}
\usepackage[most]{tcolorbox}
\usepackage{scrextend}

 \aclfinalcopy 

\title{Exploration and Discovery of the COVID-19 Literature \\ through Semantic Visualization}

\author{
Jingxuan Tu\textsuperscript{\textnormal{1}}, 
Marc Verhagen\textsuperscript{\textnormal{1}}, 
Brent Cochran\textsuperscript{\textnormal{2}}, 
James Pustejovsky\textsuperscript{\textnormal{1}}\\
  \textsuperscript{1}Brandeis University
  \textsuperscript{2}Tufts University
  \\
  \texttt{\{jxtu,verhagen,jamesp\}@brandeis.edu}\\ \texttt{brent.cochran@tufts.edu}
}
\date{}

\begin{document}
\maketitle

\begin{abstract}
We are developing semantic visualization techniques in order to enhance exploration and enable discovery over large datasets of complex networks of relations. Semantic visualization is a method of enabling exploration and discovery over large datasets of complex networks by exploiting the semantics of the relations in them. This involves (i) NLP to extract named entities, relations and knowledge graphs from the original data; (ii) indexing the output and creating representations for all relevant entities and relations that can be visualized in many different ways, e.g., as tag clouds, heat maps, graphs, etc.; (iii) applying parameter reduction operations to the extracted relations, creating “relation containers,” or functional entities that can also be visualized using the same methods, allowing the visualization of multiple relations, partial pathways, and exploration across multiple dimensions. Our hope is that this will enable the discovery of novel inferences over relations in complex data that otherwise would go unnoticed. We have applied this to analysis of the recently released CORD-19 dataset. 
\end{abstract}

\section{Introduction}

COVID-19 is the first global pandemic within a century and has resulted in hundreds of thousands of deaths as well as severe disruption of both economic and social structures worldwide.  The need to develop vaccines, therapies, and rapid tests for this virus is urgently needed.  To facilitate the scientific and medical effort to stop this pandemic, most publishers are making full text of COVID-19 related manuscripts freely available.

To further speed the scientific exploitation of COVID-19 literature, we are developing semantic visualization techniques in order to enhance exploration and enable discovery over large datasets of complex networks of biomedical relations.  Semantic visualization allows for visualization of user-defined subsets of these relations through interactive tag clouds and heat maps. This allows researchers to get a global view of selected relationship subtypes drawn from hundreds or thousands of papers at a single glance. This allows for the ready identification of novel relationships that would typically be missed by directed keyword searches. 

Due to the short  time frame within which researchers have been able to  analyze the COVID-19 literature, there is little existing work on visualization of COVID-19 articles. \citet{scisight} has developed SciSight, a tool that can be used to visualize co-mentions of biomedical concepts such as genes, proteins and cells that are found in the articles related to COVID-19. Most visualizations, however, relate to epidemiological statistics and the effects of Covid-19 on social and health factors\footnote{{\small \url{https://www.cdc.gov/coronavirus/2019-ncov/covid-data/data-visualization.htm}}}. 

In this paper we have applied semantic visualization techniques to the analysis of the recently released Harvard INDRA Covid-19 Knowledge Network   (INDRA CKN) dataset, the Blender lab Covid knowledge graph dataset (Blender KG), and the COVID-19 Open Research Dataset (CORD-19).
Our hope is that this will enable the discovery of novel inferences over relations in complex data that otherwise would go unnoticed. We have released our ready-to-use Covid-19 literature visualization web application to the community\footnote{\url{https://www.semviz.org/}}.

\begin{figure*}[]
    \centering
    \includegraphics[width=\textwidth, height=7.5cm]{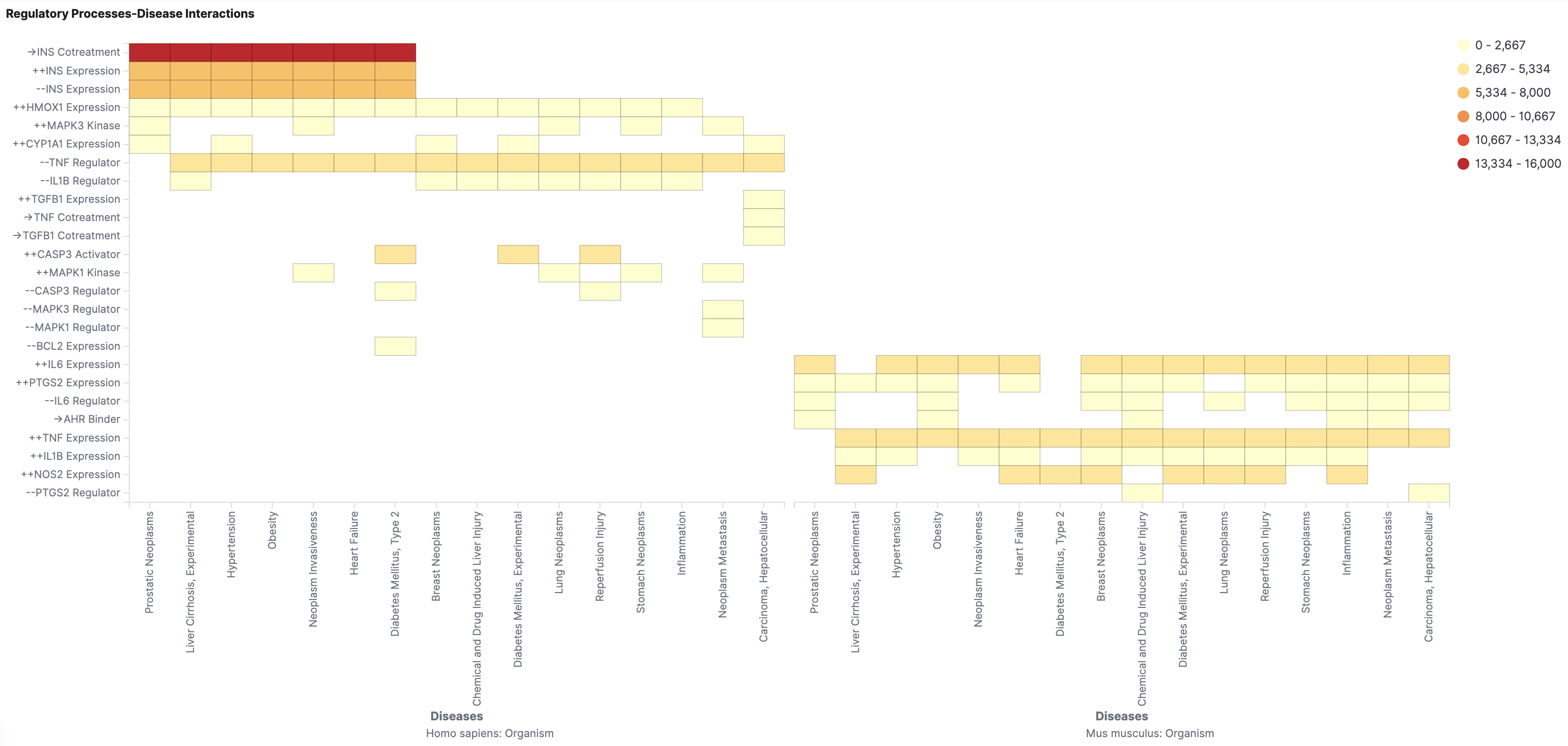}
    \caption{Regulatory Processes-Disease Interactions Heat Map}
    \label{fig:regulatoryHeatmap}
\end{figure*}

\section{Methods}

\subsection{Semantic Visualization}

 \textit{Semantic visualization} is a method of enabling exploration and discovery over large datasets of complex networks by exploiting the semantics of the relations in them. This involves (i) NLP to extract named entities, relations and knowledge graphs from the original data; (ii) indexing the output and creating representations for all relevant entities and relations that can be visualized in many different ways, e.g., such as tag clouds, heat maps, graphs, etc.; (iii) applying  parameter reduction operation  to the extracted relations, creating {\it functional types} that can also be visualized using the same methods,
 allowing the visualization of multiple relations, partial pathways, and exploration across multiple dimensions.

While individual relations derived with NLP tools can be visualized through heat maps, demonstrating how multiple relations relate to each other when chained together can be tricky to visualize, requiring cumbersome network visualization techniques  \citep{mercatelli2020gene,nelson2019embed}.
The large number of nodes and connections along with the heterogeneity of both  
node types (proteins, chemicals, diseases) and edges (structural, functional, and causal interactions) complicates the visualization \citep{agapito2013visualization,salazar2014ppi,baryshnikova2016systematic}.

For this reason, we have developed a method of parameter reduction, which reduces relations to functional types, allowing them to be treated as individuals, and visualized as tag clouds. Similarly, the arguments in a heat map relation can each refer to a functional type, enabling a chain of relations to be expressed in a conventional heat map visualization.  
This effectively results in  the creation of {\it dense tag clouds} and {\it dense heat maps}.  Figure \ref{fig:regulatoryHeatmap} illustrates such a dense heat map in the Blender KG dataset, where a functionally typed protein is implicated in a disease relation (e.g., ``those chemicals that are down regulators of TNF which are implicated in obesity'')\footnote{We use the following symbols to indicate the ``action'' in each relation: ``$++$'' = increase, ``$--$'' = decrease, ``$\rightarrow$'' = affect.}.

\begin{table*}[]
\centering

\begin{tcolorbox}[title=\textsc{\textsc{CKN dataset}}]
Evidencing sentence:
\begin{addmargin}[1em]{0em}
Alemtuzumab, \textbf{ocrelizumab\textsubscript{\texttt{[Protein]}}}, rituximab, and Cladribine may \textbf{increase the risk of acquiring and severity\textsubscript{\texttt{[Relation]}}} of \textbf{COVID-19\textsubscript{\texttt{[Protein]}}}.
\end{addmargin}
\vskip 0.1in
Raw protein-protein relation tuple: 
\begin{addmargin}[1em]{0em}
\textit{(ocrelizumab, COVID-19, Activation)}
\end{addmargin}
\end{tcolorbox}

\begin{tcolorbox}[title=\textsc{Blender KG}]
Evidencing sentence:
\begin{addmargin}[1em]{0em}
\textbf{10074-G5\textsubscript{\texttt{[Chemical]}}} results in \textbf{decreased expression\textsubscript{\texttt{[Relation]}}} of \textbf{MYC\textsubscript{\texttt{[Gene]}}} protein.
\end{addmargin}
\vskip 0.1in
Raw chemical-gene relation tuple: 
\begin{addmargin}[1em]{0em}
\textit{(10074-G5, MYC, Decrease Expression)}
\end{addmargin}
\end{tcolorbox}

\caption{Example data from PPCA and Blender KG, including extracted relations.}
\label{relation-example}
\end{table*}

\subsection{Data and Tools}

Our COVID-19 literature visualization uses information from three datasets. The COVID-19 Open Research Dataset (CORD-19)\footnote{The release date of the dateset we use is 2020-5-12. Download from \url{https://www.semanticscholar.org/cord19/download}} \citep{cord19} contains over 60,000 scientific papers on COVID-19 and related historical coronavirus research. In our work, we extract the following metadata for each article in CORD-19: \texttt{PMID}, \texttt{Title}, \texttt{Abstract}, \texttt{Authors}, \texttt{Publish time} and \texttt{Journal}.

The Harvard INDRA CORD-19 causal assertions dataset \footnote{\url{https://emmaa.indra.bio/all_statements/covid19}} contains over 200,000 causal assertions (CAs) extracted from the full text of 32,040 CORD-19 articles by multiple machine reading systems including REACH \citep{reach} and Sparser \citep{sparser}. Extracted events were assembled by the Integrated Network and Dynamical Reasoning Assembler (INDRA)\footnote{\url{https://github.com/sorgerlab/indra}} \citep{indra}. 24 relation types were defined including \texttt{Activation}, \texttt{Complex}, \texttt{Phosphorylation}, etc.  
Other than the terms already been embedded in PPIs, we also extract biomedical named entities from the CKN dataset. We apply the ScispaCy NER model\footnote{\url{https://allenai.github.io/scispacy/}} \citep{scispacy} trained on the BIONLP13CG corpus \citep{bionlp13cg} on the evidencing sentences to extract biomedical named entities. The NER model defines 14 fine-grained entity types including \texttt{Pathological Process}, \texttt{Cellular Component}, \texttt{Gene} \texttt{Chemical} and \texttt{ALL}\footnote{The annotated entity type names from the model are \texttt{PATHOLOGICAL\_FORMATION}, \texttt{CELLULAR\_COMPONENT}, \texttt{GENE\_OR\_GENE\_PRODUCT} and \texttt{SIMPLE\_CHEMICAL}. \texttt{ALL} stands for all types.}.

The Blender lab Covid Knowledge Graphs (Blender KG)\footnote{\url{http://blender.cs.illinois.edu/covid19/}} contain knowledge including entities, relations, and events that are extracted from the CORD-19 dataset through deep learning methods \cite{lin2020joint}. We use the chemical-gene, chemical-disease, gene-disease relation extraction results from Blender KG. It has over 1,640,000 relations with evidencing sentences from biomedical articles that follow the ontology introduced in the Comparative Toxicogenomics Database, and there are 13 chemical-gene relation types.

Table~\ref{relation-example} shows data samples from both the INDRA CKN and Blender KG. Each sample contains a biomedical relation and a corresponding evidencing sentence of that relation.

Our back-end data index is built with Elasticsearch, and the visualization is presented via Kibana. The Elasticsearch+Kibana stack (ELK)\footnote{\url{https://www.elastic.co/elastic-stack}} can provide text-based search and navigation functionalities that enable us to explore the text using human language technologies.

\subsection{Index Structure}

We index each dataset following a hierarchical representation. Based on the content of each dataset, we build the data indices that compose of different generic layers.

\paragraph{\textbf{Type-level layer}} Represents data that are ``entities'' or can be grounded as ``functional types''. For example, in the CORD-19 metadata, an article author name \texttt{Jane Doe} can be seen as a single entity. INDRA CKN and Blender KG both contain biomedical relations, and components such as \texttt{COVID-19} and \texttt{MYC} that are involved in the relation, as shown in Table~\ref{relation-example}, can also been seen as entities. In addition to the individual terms that have been encoded in the datasets, 
the argument and predicate of a relation can be grounded as a functional type. For example in Table~\ref{relation-example}, from the causal assertion \textit{(ocrelizumab, COVID-19, Activation)}, the entity \texttt{COVID-19 Activator} can be generated. Subsequently, it is also implied that \textit{ocrelizumab} is included in the \texttt{COVID-19 Activators} set.

\paragraph{\textbf{Phrase-level layer}} Represents data that can be transformed into ``term tuples''. A term tuple can be a natural relation tuple that is identified in the datasets. In Table~\ref{relation-example}, the relation \textit{(10074-G5, MYC, Decrease Expression)} is a tuple consisting of three terms. We can also build term tuples from indexed entities and functional types. Term tuple  \textit{(COVID-19, Viruses)} contains the entity \texttt{COVID-19} that appears in the abstract of an article, and \texttt{Viruses} is the journal name where this article is from, which can also be indexed as an entity. Building such relation tuples allows deeper relationship representation and visualization.

\begin{figure*}[ht]
        \centering
        \fbox{\includegraphics[width=\textwidth, height=7.8cm]{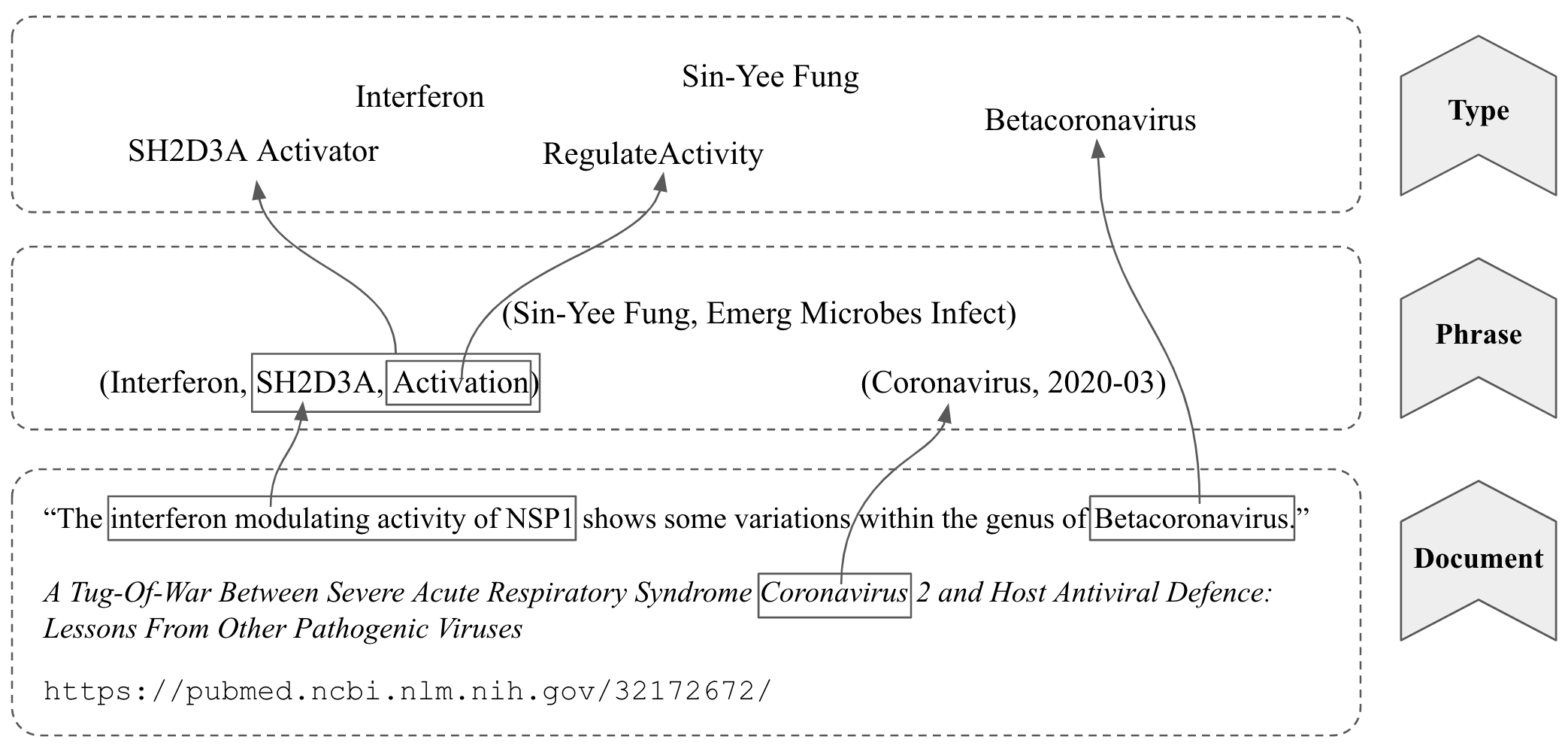}}
        \caption{Hierarchical index representation for the datasets. Boxes from bottom to top show how data is represented in different layers. Arrows show how data is passed and decomposed between layers.}
        \label{index-struct}
\end{figure*}
    
\paragraph{\textbf{Document-level layer}} Represents data as documents that provide context information to the indexed functional entities and term tuples. The document text is of various length and it can be a phrase, sentence, or a whole paragraph. Primarily, we index evidencing sentences, article titles and abstracts as documents. A clickable PubMed URL is also indexed to show the provenance of each evidencing sentence and article title.

Figure~\ref{index-struct} gives an example of the data that is processed into the hierarchical index representation. At the document level we can have the evidencing sentence, article title and the URL link to the full text. Some extracted relations and entities can be fitted into the other two layers. 
For example, \texttt{Betacoronavirus} is an entity of protein; \texttt{Coronavirus} can be used to form a new term tuple with \texttt{2020-03} of type \textit{(keyword in abstract, Publish time)}. Terms in a term tuple can also directly come from the dataset. In the phrase-level layer shown in Figure~\ref{index-struct}, the author name \texttt{Sin-Yee Fung} and journal name \texttt{Emerg Microbes Infect} are from the CORD-19 dataset. Terms from the phrase-level layer can become entities or have associations with newly generated entities in the type-level layer. In Figure~\ref{index-struct}, we generate the entity \texttt{RegulateActivity} and the functional type \texttt{SH2D3A Activator} in the type-level layer, and they are all associated with the tuple \textit{(Interferon, SH2D3A, Activation)} in the phrase level. \texttt{RegulateActivity} is the parent relation of \texttt{Activation}\footnote{We followed the biological relationship hierarchy identified in \url{https://indra.readthedocs.io/en/latest/modules/statements.html\#}}. 
The protein \texttt{Interferon} can be grounded as an \texttt{SH2D3A Activator}\footnote{\texttt{SH2D3A} is an alias of \texttt{NSP1} in the evidencing sentence.}.

In the implementation, data indices are stored in the JSON format after processing. Data in the different layers from our hierarchical index representation is presented using different visualizations.

\subsection{Visualization Types}

Kibana supports various types of visualization panes that allow us to explore the indexed data. We choose the combination of panes that are compatible with our hierarchical data index and allow us to design and build semantically meaningful interactive visualization strategies.

\paragraph{\textbf{Tag Cloud}} displays data as a group of words. The size of each word reflects its importance. This pane can provide us with a clear view of the essentials we want to explore, such as popular proteins involved in different relations or author names that make substantial contributions. We use tag clouds to display entities and functional types that can be categorized in the same group based on our standards.

\begin{figure}[ht]
    \centering
    \fbox{\includegraphics[width=0.95\columnwidth, height=4.7cm]{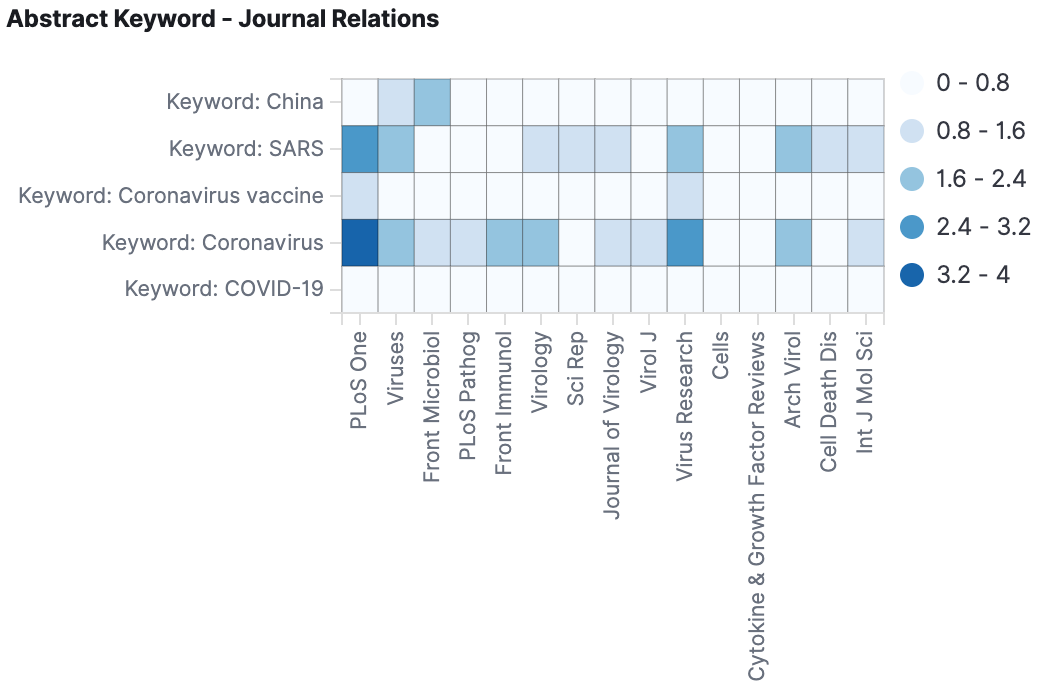}}
    \caption{Sample Heat Map.}
    \label{sample-heatmap}
\end{figure}

\paragraph{\textbf{Heat Map}} displays data as a grid matrix. The color shade of each cell in the matrix reflects its importance. Heat maps can take advantage of color density to show data clusters in two dimensions, so we use the heat map to display term pairs. In Figure~\ref{sample-heatmap}, the sample heat map pane represents the relations between selected entities in the article abstract and journal names. The intersected cells between the entity \texttt{Coronavirus}, \texttt{SARS} and journal names such as \texttt{PloS One}, \texttt{Virus Research} are darker than other cells, showing that these term tuples show up more frequently.

\begin{figure*}[ht]
        \centering
        \fbox{\includegraphics[width=0.95\textwidth, height=4cm]{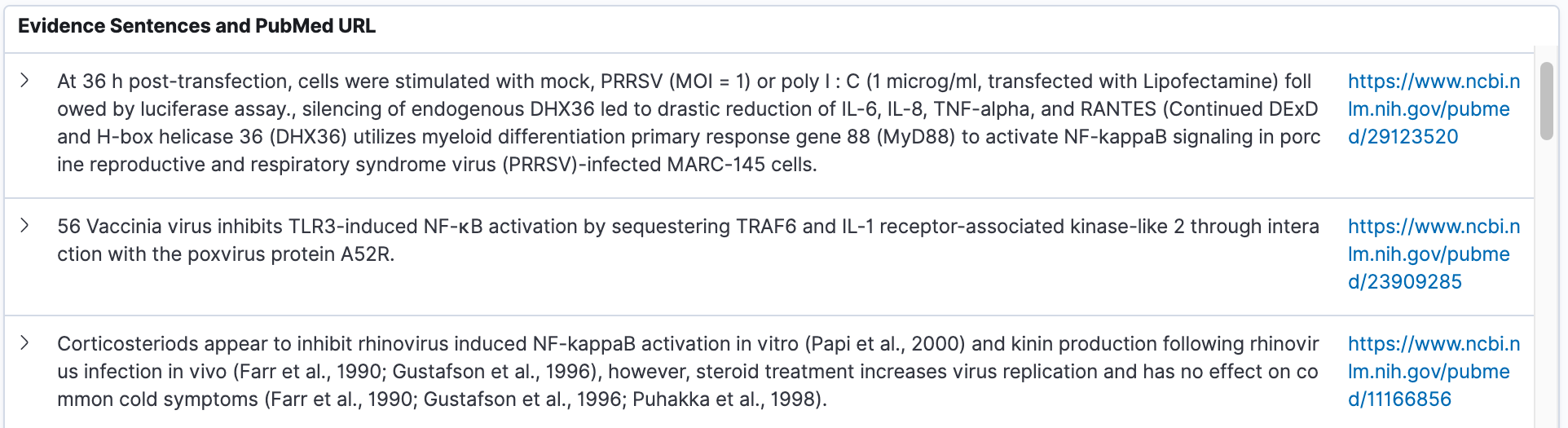}}
        \caption{Sample Data Table.}
        \label{sample-datatable}
\end{figure*}

\paragraph{\textbf{Data Table}} is flexible enough to represent any data as a table. It can have multiple columns and rows to show different data types and values. We use data tables to present document data in our index, since this visualization pane is suitable to show free text and miscellaneous information in a neat and structured way. The sample data table in Figure~\ref{sample-datatable} shows the evidencing sentences of biomedical relations and corresponding PubMed URLs that link to the full article. Each row in the data table can also be expanded to show more additional information that has been processed and indexed.

\paragraph{\textbf{Metric}} is a calculation as a single number that can be used to show simple statistics computed from the data index. We use it to show the number of PubMed URLs/articles that have been indexed. It can provide an overview of the whole dataset.

\paragraph{\textbf{Bar Chart}} uses bars to represent the continuous values of data along different axes. It can be used to represent term tuples.

\section{COVID-19 Literature Visualization}

Kibana dashboards are collections of visualizations, searches and maps. Dashboards elements can be arranged as desired and visualizations will be updated in real-time when a search is performed. Dashboards provide quick insights into the underlying data and enable you to drill down into details. Given the data indices we built for the three datasets, we create two Kibana dashboards. The Covid CA dashboard holds visualizations designed principally for protein-protein relations from the CKN dataset, and the Covid KGs dashboard contains visualizations designed for the Covid-19 knowledge graphs from the Blender KG dataset.

Dashboards are fulfilled as web applications. The navigation of a dashboard is mainly through clicking and searching. By clicking the protein keyword \texttt{EIF2AK2} in the tag cloud named ``Enzyme proteins participating Modification relations'', a constraint on the type of proteins in modifications is added. Correspondingly, all the other visualizations will be changed. For example, the ``Evidence Sentences and PubMed URL'' data table will display evidencing sentences that only involve \texttt{EIF2AK2} in the relations. The ``Abstract Keyword - Journal Relations'' heat map will form new color shade clusters based on the new set of articles that mentioned \texttt{EIF2AK2}. One can also put a query into the search box to navigate the dashboard. More detailed interaction and navigation are discussed in Subsections~\ref{ppi-viz} and \ref{kg-viz}.

\begin{figure}[]
    \fbox{\includegraphics[width=0.9\columnwidth, height=5cm]{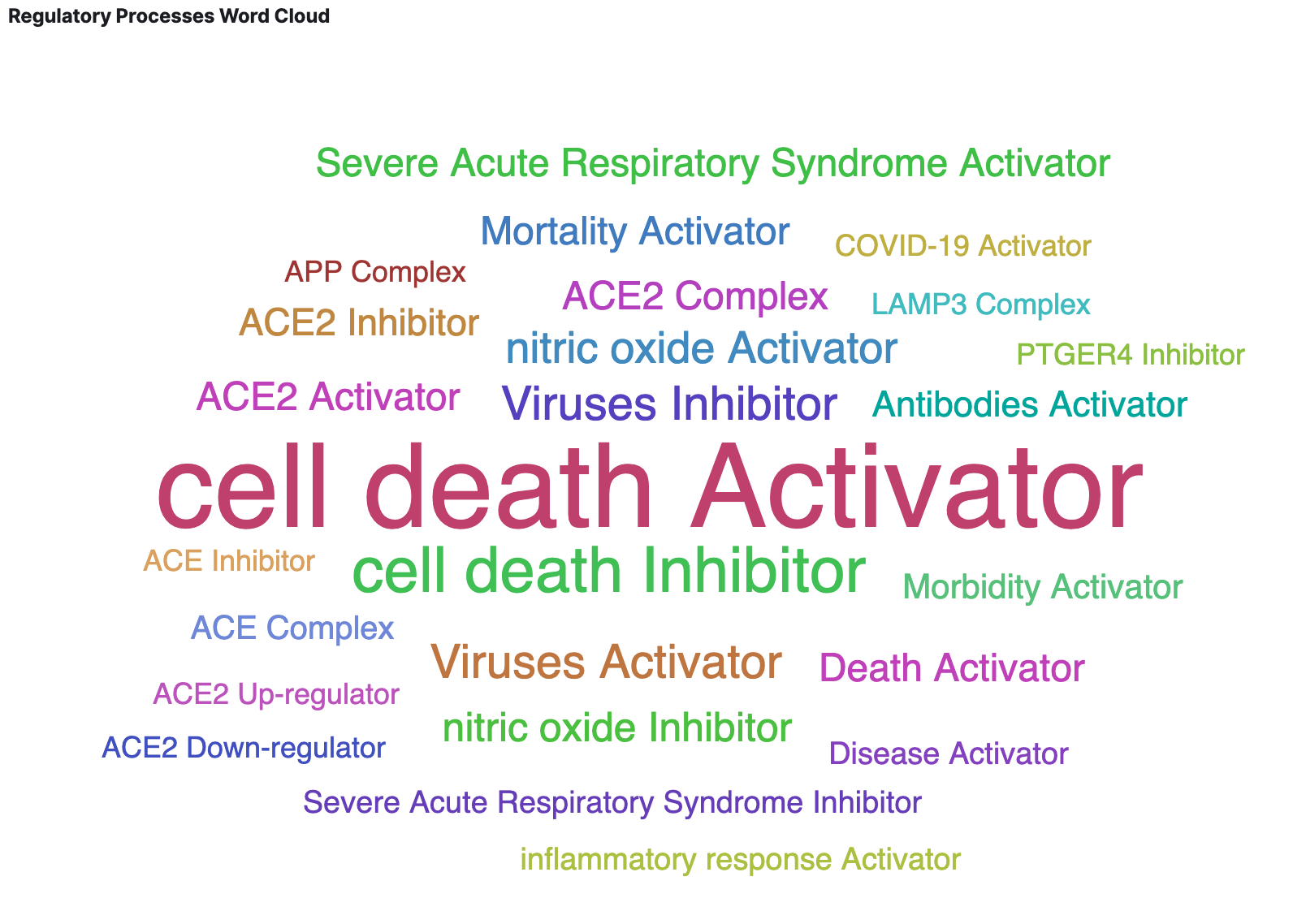}}
    \caption{Visualization of protein regulators}
    \label{sample-regulators}
\end{figure}

We did not build a separate dashboard for the CORD-19 dataset. But note that CORD-19 was used to add visualizations of metadata that are included in the other two dashboards. In practice, we take the \texttt{PubMed ID}, \texttt{Title}, \texttt{Abstract}, \texttt{Authors}, \texttt{Publish time} and \texttt{Journal} of each article as the metadata from CORD-19. Then we use the \texttt{PubMed ID} as the unique identifier to align the metadata of each article with the evidencing sentence where the relations are extracted from.

\subsection{The Covid Causal Assertions Visualization}
\label{ppi-viz}

The Covid CA dashboard contains a set of visualizations that are designed to enable users to discover the novel inferences of protein-protein interactions and associated context information. 
Since visualizations are built on the indexed data, users can type in a query to search for relevant CA and context information. We include several kinds of visualizations: (1) data tables for tracing evidence associated with relations, (2) metrics panes to display the count of evidences and the count of unique articles, (3) tag clouds and heat maps for some metadata, (4) type-level and phrase-level visualizations that enable users to drill down into the elements in the relations, (5) dense visualizations for functional types, and (6) visualizations of upstream regulators. We now elaborate on the last three of these.

{\em Type-level and phrase-level visualizations}. Each CA contains three elements: protein-A, relation type, and protein-B. We group the 24 relation types into two ``metatypes'': \texttt{RegulateActivity} and \texttt{Modification}. Furthermore, protein-A and protein-B involved in RegulateActivity relations are categorized into \texttt{Subject} and \texttt{Object}. Protein-A and protein-B involved in Modification relations are categorized into \texttt{Enzyme} and \texttt{Substrate}. We believe this categorization allows our visualizations to conform to biological convention. On the dashboard, we create tag clouds for these categories. We also create a subject-object interaction heat map to show regulatory relationships, an enzyme-substrate interaction heat map to show protein modification relationships, and heat maps for some common relation types such as \texttt{Activation} and \texttt{Inhibition}. Finally, we include tag clouds for entity types extracted with the ScispaCy NER model.

{\em Visualizations for functional types}. As mentioned in Section 2.1, we also enable the visualization of CAs by applying parameter reduction, which is a critical step in semantic visualization. Given two CA tuples \textit{(Protein-A, Activation, Protein-B)} and \textit{(Protein-B, Activation, Protein-C)}, we create the functional type {\em Protein-C-Activator} with members {\em Protein-A} and {\em Protein-B}. We can now have a tag cloud for all functional types (see Figure~\ref{sample-regulators}) and a separate tag cloud for the subject proteins associated with them. Clicking one of the functional types will restrict the subject proteins to just the ones involved in the functional type selected. 

{\em Visualizations for upstream regulators}. We define two types of second order CAs: one that has the same relation type as the functional type, and one that has the opposite relation type. In the dashboard, we add the ``Upstream Regulators'' tag cloud and the ``Opposite Upstream Regulators'' tag cloud to display second order relations. For example, with a functional type {\em Interferon-Activator} the "Upstream Regulators" tag cloud would include all proteins X that activate one of the Interferon activators, thereby generating a novel inference from {\em X} to {\em Interferon}. Through navigation over the keywords in each tag cloud, one can easily check the evidencing sentences of deeper CAs that are inferred through parameter reduction. 


\subsection{The Covid KGs Visualization}
\label{kg-viz}

The Covid KG dashboard contains a collection of visualizations that enable the discovery of the relationships among genes, chemicals and diseases that are related to the COVID-19. The Blender KG has a number of chemical-gene, chemical-disease and gene-disease relations and these relations are supported by the evidencing sentences not only from COVID-19 articles but also from other various medical articles. Thus, the most challenging part in the visualization is to simplify and unify the complex relations while displaying the information in breadth and depth.

We start by making the connections between chemical-gene and gene-disease relations using the same gene entries that appear in both sides.
Then we index the new chemical-gene-disease relations and visualize it via chemical-gene sub-relation heat map and gen-disease sub-relation heat map. These two heat maps are designed to be interactive with each other to show the full chemical-gene-disease triplet relations, as well as to be flexible enough to be controlled by enabling or disabling arguments of the triplet relations. 

Similar to the Covid CA dashboard, we build a data table that displays evidencing sentences and PubMed URLs, as well as tag clouds of chemicals, genes and diseases coming from the relations. Users can navigate the dashboard to find relevant context information by filtering on entities from the tag clouds. 
 we also create a tag cloud of gene semantic types by grounding chemical-gene relations. For example, given a chemical-gene term tuple (D014013, Decrease Reaction, CASP3), the semantic type \texttt{--CASP3 Regulator} is generated. 

\section{Example Use Case Scenarios}
 
This section presents recent user comments from biologists who have interacted with the SemViz platform in their own research on coronaviruses. We present three use case scenarios, each highlighting different aspects of the navigation and exploration capabilities provided by system. 

\paragraph{Link between ACE2 receptors and cytokine activation}

Prof. Lakshmi Pulakat (Tufts Medical Center)
is interested in exploring the linkage between angiotensin II type 2 receptor AT2R, IL-6 inhibition and general cytokine activation. A search in SemViz shows a linkage between these terms and respiratory distress, a key cause of hospitalization and mortality in COVID-19. The linked literature shows that AT1R inhibition significantly decreases the amount of IL-6; however, this is not a viable therapeutic strategy in patients with normal AT1R expression and normal blood pressure, as inhibition of this receptor lowers blood pressure.  Moreover,  AT1R blockade is contra-indicated in pregnancy. AT2R activation may have a similar effect on IL-6 levels without impacting blood pressure. This is a new interaction that she had previously not known, and is one that she will explore in her research.

\paragraph{Regulation of Interferon by coronaviruses}

Prof. Marta Gaglia (Tufts University School of Medicine, Dept. of Microbiology) is interested in mining existing literature for interactions between proteins from all coronaviruses and the type I IFN system, in order to generate hypotheses on the function of IFN-regulating proteins in SARS-CoV-2 infection. Using a search for coronavirus and interferon, and COVID and interferon, she pulled out a lot of information on existing links between SARS-CoV-2 and other coronaviruses and the IFN induction pathway. While she was previously aware of some of the information, she also identified several new links to follow up on. For example, one of the high search tags that were pulled out was glycosylation of the coronavirus M protein and its role in IFN regulation in different coronaviruses. Another connection, DDX58 inhibitor and NOP53, pointed to the role of a nucleolar protein in IFN regulation during coronavirus infection. Whether similar regulation occurs in SARS-CoV-2 will be worth testing. In addition, she found links to a number of publications that were more distantly related to the topic that she had missed in previous searches that proved useful in the design of the project that she is currently proposing.

\paragraph{Proteases as therapeutic targets}

Brent Cochran (Tufts University, Dept. of Developmental, Molecular, and Chemical Biology) explored his interest in cellular proteases as therapeutic targets for COVID-19.  He did a search in SemViz for COVID-19 proteases.  In the regulatory tag cloud, TMPRSS2 was the most frequent regulator as expected since it is known to be required for viral fusion. However, TMPRSS4 also showed up in the tag cloud as a regulator.  Clicking on it brought up the evidence in influenza that both TMPRSS2 and TMPRSS4 can cleave the viral fusion protein. This raises the question whether the same is true for COVID-19 and whether we need to consider the expression and inhibition of both proteases to inhibit the virus through this route.  This is something new that he did not previously know that was identified by the visualization.

\vspace*{-2mm}
\section{Exploring Pathways}
\vspace*{-2mm}
Kibana allows compact visualization of functional types and relations between elements. But it is not well-suited for a flexible visualization of pathways because the underlying index is a document database where a graph database would be more appropriate. We have experimented with a simple browser implemented on top of a graph where in principle we have access to all regulation pathways between proteins. Since full pathway construction from relations is computationally intractable we limit the length of the pathways we are constructing. We do this both by imposing a maximum length of 5 as well as by estimating the number of upstream regulators we may get for each functional type,  reducing the length of the pathway accordingly if the estimate is too high. Users can search for a protein name and will be presented with a functional type (at the moment only Activation is used), listing both the members of the set (i.e., the activators of the protein) and upstream regulators.

\begin{figure}[ht]
    \centering
    \fbox{\includegraphics[width=6cm]{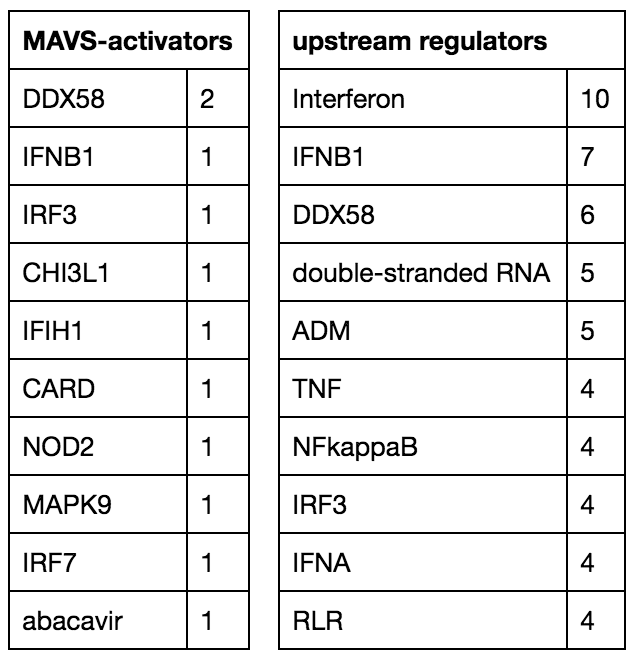}}
    \caption{MAVS Activators and upstream regulators in the Pathways Browser}
    \label{mavs-regulators}
\end{figure}
\begin{figure}[ht]
    \centering
    \fbox{\includegraphics[width=7.3cm]{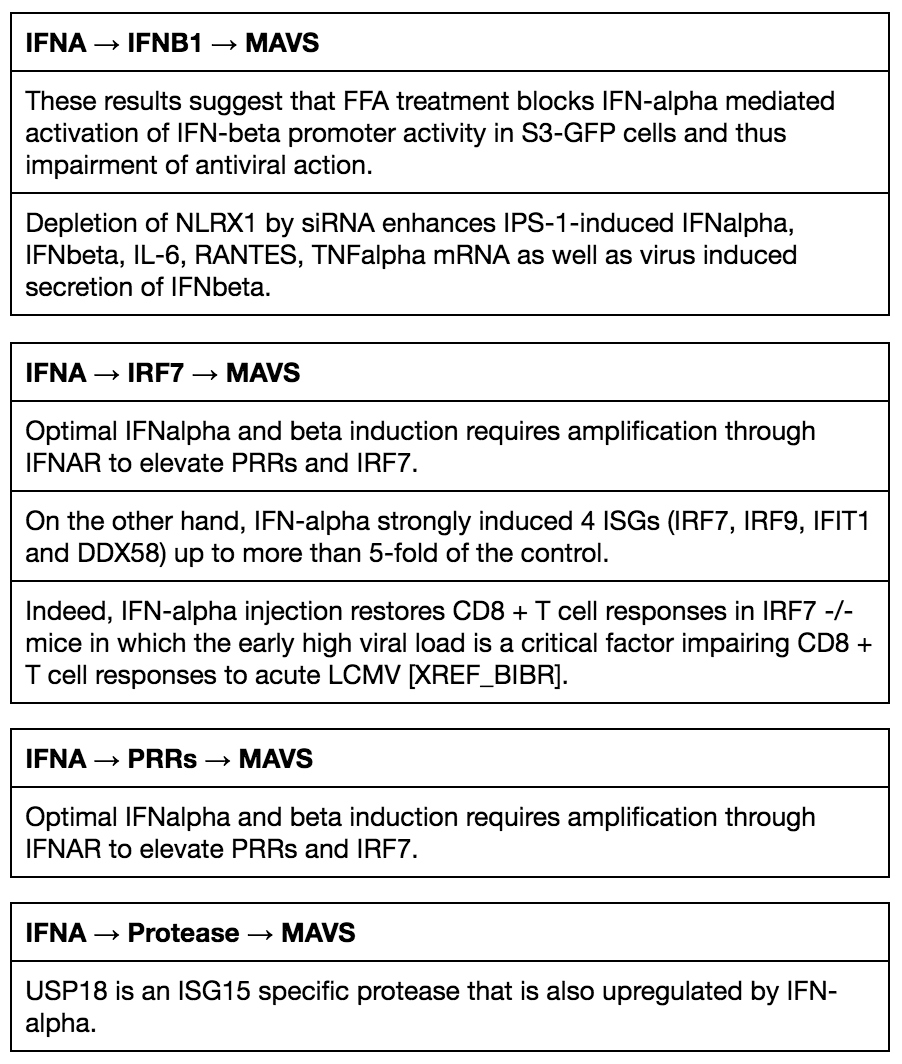}}
    \caption{Displaying evidence for upstream regulators in the Pathways Browser}
    \label{mavs-ifna}
\end{figure}

\vspace*{-2mm}
In Figure \ref{mavs-regulators} we show a partial result of a search for MAVS, with the 10 most common activators of MAVS and the 10 most common proteins that activate one of the MAVS activators (so here the length of the pathway is set to 3). From the list of activators we can select one to see all the evidence presented, with links to the sources. Similarly, we can explore the evidence for upstream regulators. Displayed in Figure \ref{mavs-ifna} is the evidence that would be presented if we select the IFNA upstream regulator in right column of Figure \ref{mavs-regulators}. Individual pathways starting with IFNA and ending with MAVS are presented with evidence for the first regulatory relation in the pathway, and the evidence again can be followed to the source. 

\vspace*{-6mm}
\section{Future Work and Conclusion}

In this paper, we have presented a platform for the semantic visualization of multiple types of entities and relations extracted from the CORD-19 dataset. This involves indexing the output of  NLP readers in  several pipelines, in order to give semantically typed elements for tag clouds and heat maps. A unique technique developed here is the application of  parameter reduction operations to the extracted relations, creating “relation containers,” or functional entities that can also be visualized using the same methods, allowing the visualization of multiple relations, and both  partial and complete protein-protein pathways. We are currently working to provide graph-based visualizations of network relations that go beyond the heat map capabilities provided in Kibana-like dashboards. 
It is our hope that this visualization environment  will enable the discovery of novel inferences over relations in complex data that otherwise would go unnoticed.

\subsubsection*{Acknowledgements} 

This work is supported in part by US Defense Advanced Research Projects Agency (DARPA), Contract W911NF-15-C-0238; and  DTRA grant DTRA-16-1-0002; Approved for Public Release, Distribution Unlimited. The views expressed are those of the authors and do not reflect the official policy or position of the Department of Defense or the U.S. Government.   All remaining errors are, of course, those of the authors alone. 


\bibliography{acl2020}
\bibliographystyle{acl_natbib}

\end{document}